\documentclass[letterpaper, 10pt, conference]{ieeeconf}  

\IEEEoverridecommandlockouts                              
\usepackage{cite}
\usepackage{amsmath,amssymb,amsfonts}
\usepackage{algorithmic}
\usepackage{graphicx}
\usepackage{textcomp}
\usepackage{xcolor}
\usepackage{algorithm}
\usepackage{url}
\def\BibTeX{{\rm B\kern-.05em{\sc i\kern-.025em b}\kern-.08em
    T\kern-.1667em\lower.7ex\hbox{E}\kern-.125emX}}

\graphicspath{ {img/} }

\DeclareMathOperator{\e}{e} 

\begin{document}

\title{Improving Learning Effectiveness For Object Detection and Classification in Cluttered Backgrounds
}

\author{Vinorth Varatharasan$^{1}$, Hyo-Sang Shin$^{1}$, Antonios Tsourdos$^{1}$ 
\thanks{$^{1}$ School of Aerospace, Transport and Manufacturing, Cranfield University, Cranfield MK43 0AL, UK
(email: \texttt{vinorth.varatharasan@gmail.com})} and Nick Colosimo $^{2}$ \thanks{$^{2}$ BAE Systems Warton Aerodrome, Warton, Preston, PR4 1AX}
}

\maketitle

\begin{abstract}

Usually, Neural Networks models are trained with a large dataset of images in homogeneous backgrounds. The issue is that the performance of the network models trained could be significantly degraded in a complex and heterogeneous environment. To mitigate the issue, this paper develops a framework that permits to autonomously generate a training dataset in heterogeneous cluttered backgrounds. It is clear that the learning effectiveness of the proposed framework should be improved in complex and heterogeneous environments, compared with the ones with the typical dataset. In our framework, a state-of-the-art image segmentation technique called DeepLab is used to extract objects of interest from a picture and Chroma-key technique is then used to merge the extracted objects of interest into specific heterogeneous backgrounds. The performance of the proposed framework is investigated through empirical tests and compared with that of the model trained with the COCO dataset. The results show that the proposed framework outperforms the model compared. This implies that the learning effectiveness of the framework developed is superior to the models with the typical dataset.

\end{abstract}

\section{INTRODUCTION}

There have been growing attention on the exploitation of potential benefits of Unmanned aerial vehicles (UAVs). Their potential applications under discussion are extensive, e.g., surveillance, delivery, defence, agriculture, and so on. For instance, if a small UAV is considered, it can be used in civil applications such as delivery or supply, or in defence applications such as military resupply or sacrificial weapons; therefore a trade-off between a low cost and good performance is needed. Many sensors exist, such as lidars, radars, electro-optical cameras, etc. The major challenge of the latter is that there is not direct range information since there are difficulties to differentiate between a close and a small object, or a far and a big object, because the angle subtended is the same.

This is the reason behind the use of neural networks in parallel. These tools are powerful for computer vision applications and can be very useful in sense-and-avoid or other applications. For example, if a particular kind of object is classified, stadiametric rangefinding and simple trigonometry could be used to find the depth of an object. 

The close and frequent proximity to urban obstacles and numerous other airspace users (UAV in particular) presents a new level of challenge. However, there are also other applications such as situational awareness, in which robust detection algorithms have to be implemented. 

Last but not least, background clutter easily confuses the machine. Indeed, when the hardware is running object detection and tracking algorithms, the results can be false because of the similarity between the object of interest (target) and the background.

Existing state-of-the-art object detection and tracking techniques provide great performances when observing the objects of interest with homogeneous backgrounds. However, the performance dramatically degrades when observing the objects of interest with cluttered backgrounds. The main aims are thus to have good performances in any background, and for that, an innovative method of training is introduced. Therefore, the following objectives were considered:

\begin{itemize}
	\item Improve detection and classification probability while minimising false alarm rate which is a function of the background scene clutter.
	\item Verify and validate the proposed methodology by using existing detection and classification accuracy metrics, by comparing the performances in heterogeneous backgrounds.
\end{itemize}

\vskip 3pt

The rationale behind the poor performance of the state-of-the-art object detection techniques in cluttered backgrounds is the lack of the use of heterogeneous backgrounds in the training set. 

\vskip 7pt

Thus, instead of using homogeneous backgrounds which are typically used in training, heterogeneous backgrounds are introduced in the training dataset by creating an artificial data using Deep Learning techniques to improve the learning efficiency. This proposed framework is automated to create a large training dataset, and hence help achieve the requirements by obtaining accuracy improvements in object detection. 

\vskip 7pt

Nowadays, more and more background replacement tools using Artificial Intelligence are being implemented in various sectors. For instance, Microsoft develops this for Skype for Business~\cite{Microsoft-research}, and Google AI tries to remove background from YouTube videos in real-time~\cite{Google-research} using Chroma-keying. Many other open-source tools are also available to the public.

\vskip 7pt

Some research has also been done to obtain better performance in cluttered scenes (Feature extraction~\cite{EXTRACTION1}~\cite{EXTRACTION2}~\cite{FEATURE-EXTRACTION}, Filtering~\cite{TEMPORAL}, and so on). Furthermore, data augmentation is widely used today in order to improve the learning effectiveness, for example by changing the lightning, the rotation, the size, the variety of backgrounds (e.g. snow, rain, etc.) of the images. A similar approach to our proposed methodology, explained in Chapter~\ref{chap2}, was introduced in~\cite{SIMILAR-APPROACH}. This study also tried to improve training effectiveness for object classification in cluttered environments, but without using Deep Learning techniques. In fact, background removal must have been done manually and spectral texture-based features were used for back-propagation. Compared to the former approaches, the use of AI techniques generates a dataset with better accuracy and in a very limited time.

\vskip 7pt

The paper starts with the explanation of the proposed framework, which is split into two essential parts: an object segmentation technique performing a background removal and a chroma-keying process conducting a background replacement. These two techniques are used to generate autonomously a dataset of images in cluttered backgrounds which is then used to train the object detection model. The performance assessment is then considered to validate the proposed approach by using numerical simulations, with the comparison between a model trained with the COCO dataset and the model trained with the dataset generated by the proposed framework. Finally, the main outcomes are summarised, followed by critics and proposals.

\section{PROPOSED METHODOLOGY}

\label{chap2}

The integration of two different techniques is used to respect the aims and objectives set in the Introduction:
\begin{itemize}
	\item An image segmentation technique called DeepLab.
	\item The Chroma-keying technique.
\end{itemize} 

\vskip 3pt

In other words, these techniques are used to obtain an unreal dataset of images with heterogeneous cluttered backgrounds, which is then used to train the Neural Network model. By implementing this strategy, the learning (or training) effectiveness improves and therefore, the detection and classification accuracy gets better in heterogeneous backgrounds and by extension, in any background.

\vskip 7pt

For example, the Neural Network is trained on a large dataset in different cluttered backgrounds (e.g. $2000$ images $\times$ $5$ cluttered backgrounds = dataset of $10000$ images), which is also time-saving because only $2000$ images need to be labelled as ground truths for $10000$ images (5 times faster).

\subsection{Image segmentation}

On the one hand, the object detection generates bounding boxes around the objects of an image and classifies each one of these boxes into a class. But on the other hand, the object segmentation classifies each pixel of the image into a class, which permits to locate the detected objects/instances more precisely.

\vskip 7pt

There are two types of image segmentation:

\begin{itemize}
	\item Semantic segmentation.
	\item Instance segmentation.
\end{itemize}

\begin{figure}[!ht]
\centering
{\includegraphics[width=0.8\linewidth]{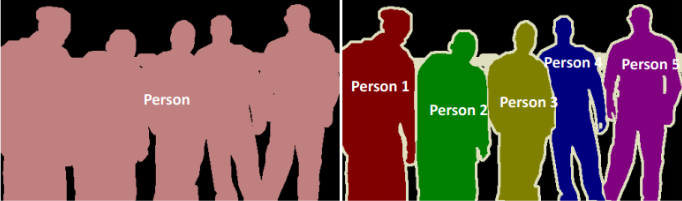}}
\caption{Semantic segmentation (left) and Instance segmentation (right)~\cite{SEGvsINST}}
\label{types_segmentation}
\end{figure}

Many state-of-the-art semantic algorithms exist:

\begin{itemize}
	\item Fully convolutional networks~\cite{FCN}.
	\item Encoder-decoder networks: SegNet~\cite{SEGNET}, U-Net~\cite{UNET}, DeconvNet~\cite{DECONVNET}.
	\item Dilated networks: DilatedNet~\cite{DILATEDCONV}, PSPNet~\cite{PSPNET}, DeepLab (v1~\cite{DEEPLABV1}, v2~\cite{DEEPLABV2}, v3~\cite{DEEPLABV3}, v3+~\cite{DEEPLABV3PLUS}).
\end{itemize}

Same for region-based instance segmentation algorithms:

\begin{itemize}
	\item Multi-task Network Cascade (MNC~\cite{MNC})
	\item Fully Convolutional Instance-aware Semantic Segmentation (FCIS~\cite{FCIS})
	\item Mask-RCNN~\cite{MASK-RCNN}
\end{itemize}

Some semantic segmentation techniques may be real-time; however, instance segmentation is much more powerful and interesting for their accuracy performances in extracting the maximum details from the objects of interest.

\vskip 7pt

Many state-of-the-art instance segmentation methods exist, and similar techniques always get improved with time. The average precision (AP) is usually used as an accuracy metric, which is explained in Section~\ref{AP_IoU_etc}.

\begin{table}[h!]
  \begin{center}
    \caption{COCO challenge results for instance segmentation techniques~\cite{MASK-RCNN}}
    \label{trade_off}
    \scalebox{0.7}{\begin{tabular}{l|l|c c c|c c c} 
          & backbone & AP & AP$_{50}$ & AP$_{75}$ & AP$_{S}$ & AP$_{M}$ & AP$_{L}$ \\
			\hline
	  MNC & ResNet-101-C4 & 24.6 & 44.3 & 24.8 & 4.7 & 25.9 & 43.6 \\
      FCIS +OHEM & ResNet-101-C5-dilated & 29.2 & 49.5 & - & 7.1 & 31.3 & 50.0\\
      FCIS+++ +OHEM & ResNet-101-C5-dilated & 33.6 & 54.5 & - & -& -& - \\
            \hline
      \textbf{Mask-RCNN} & ResNet-101-C4 & 33.1 & 54.9 & 34.8 & 12.1 & 35.6 & 51.1 \\
      \textbf{Mask-RCNN} & ResNet-101-FPN & 35.7 & 58.0 & 37.8 & 15.5 & 38.1 & 52.4\\
      \textbf{Mask-RCNN} & ResNeXt-101-FPN & \textbf{37.1} & \textbf{60.0} &  \textbf{39.4} & \textbf{16.9} & \textbf{39.9} & \textbf{53.5} \\
    \end{tabular}}
  \end{center}
\end{table}

Thus, without tricks, Mask-RCNN outperforms all existing, single-model entries on every task, even the COCO challenge winners of 2015 and 2016. As a result, \textbf{Mask-RCNN} can be used for object segmentation purposes.

\vskip 7pt

However, since only the background needs to be removed, if we only obtain the shape of the desired objects like in Figure~\ref{types_segmentation}, we can extract these objects' shapes with semantic segmentation techniques which present better accuracy than Mask-RCNN for these goals, as described in Table~\ref{tab:perf}.

\begin{table}[h!]
  \begin{center}
    \caption{Performance (\textit{mean IoU}) of some image segmentation techniques}
    \label{tab:perf}
    \begin{tabular}{|l|c|} 
		  \hline
      Model name & \textit{mIoU} \\
			\hline
			FCN & $62.2$ \\
      DeepLabv2-CRF & $79.7$ \\
      PSPNet & $85.4$ \\
      DeepLabv3 & $87.3$ \\
			DeepLabv3+ & $89.0$ \\
      Mask-RCNN & $\textbf{74.4}$ \\
			\hline
    \end{tabular}
  \end{center}
\end{table}

Hence, semantic techniques such as \textbf{DeepLab} have much better results than Mask-RCNN and can be used to meet the predefined requirements. Therefore, we use DeepLabv3 for image segmentation.

\vskip 7pt

\label{seg} 

\begin{figure}[!ht]
\centering
{\includegraphics[width=1\linewidth]{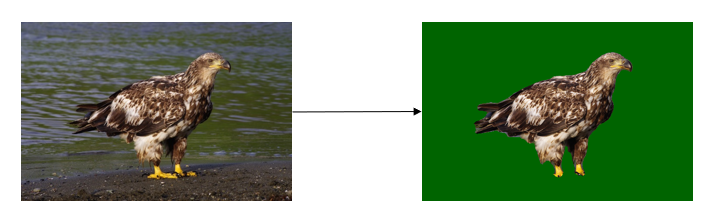}}
\caption{Object segmentation for background replacement}
\label{bg_replacement}
\end{figure}

First of all, a semantic image segmentation technique with DeepLabv3~\cite{GITDEEPLAB} in TensorFlow (Python) is introduced. It is used to remove the background of every image from the training dataset, which is then replaced by a green background, as illustrated in Figure~\ref{bg_replacement}.

The background extraction algorithm is implemented as follows:

\begin{enumerate}
	\item The latest version of the pre-trained DeepLab model is loaded. Two models are proposed: the MobileNetv2~\cite{MOBILENET} model and the Xception~\cite{XCEPTION} model. The MobileNetv2 model is much faster but less accurate than the Xception model. MobileNetv2 is a fast network structure intended for mobile devices, whereas Xception is a more robust network structure designed for server-side deployment. Thus, the latter is used since accuracy is essential for our application.
	\item After the model is loaded, the inference is run on all images in a folder named \textit{input-background-removal} so that all classes (objects) defined in the pre-trained model are segmented.
	\item Finally, a distinction between the segmented objects and the background is made. Therefore, the background can be easily removed and replaced by a green screen background by defining the RGB pixels to $[0,100,0]$.
	\item All the images with the green screen background are outputted in a folder named \textit{input-with-green-screen-background} so it can be used for the chroma-key technique in Section~\ref{chr}.
\end{enumerate}

\subsection{Chroma-key} 

\label{chr} 

\begin{figure}[!ht]
\centering
{\includegraphics[width=1\linewidth]{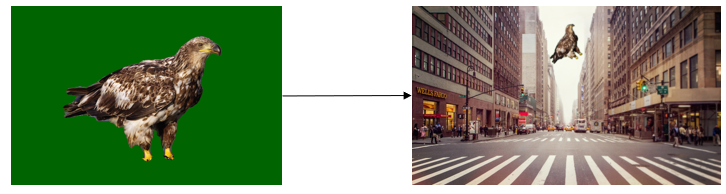}}
\caption{Chroma-key technique}
\label{chroma_key}
\end{figure}

Secondly, a Chroma-keying technique is used to remove the green background from the training dataset (e.g. the output of the first background replacement in Figure~\ref{bg_replacement}) to replace it by any cluttered background as illustrated in Figure~\ref{chroma_key}.

\newpage

The chroma-keying algorithm~\cite{GITCHROMAKEY} uses the skimage~\cite{SCIKIT} Python library to realise many imaging operations, such as:

\begin{itemize}
	\item Green background removal from an image.
	\item Merging of several pictures into one, which also permits necessary rotations, translations and scale operations of the objects of interest.
	\item Filtering of the image.
	\item Storage of the outputs.
\end{itemize}

\subsubsection{Green background removal}

The green background is removed from the input images depending on a predefined threshold.
	
\vskip 7pt
	
The \textit{NumPy}~\cite{NUMPY} library enables to use an array of four-channel image: $RGB\alpha$.

\vskip 7pt

Then, the ratio of the red/green/blue channels based on the max-bright of the pixel is obtained by dividing the RGB pixels by the norm factor (therefore $255$):

\begin{equation}
colour_{ratio} = \frac{colour_{pixel}}{255}
\label{eq:normfactor}
\end{equation}

where the colour is red, blue or green.

\vskip 7pt

Dark pixels are almost equal to zero. Consequently, when calculating the red/blue vs green ratios, we can obtain small negative values for dark pixels (the "/" sign is here used as "or"). An additional parameter of $0.2$ is added to these values to avoid this issue. 

\begin{equation}
(red/blue)vs(green) \equiv (red/blue)_{ratio} - green_{ratio} + 0.2
\label{eq:dark}
\end{equation}

The additional parameter was manually tuned to obtain better performances for the green colour removal. 

\vskip 7pt

\begin{figure}[!ht]
   \begin{minipage}{.45\linewidth}
			\includegraphics[width=1.\linewidth]{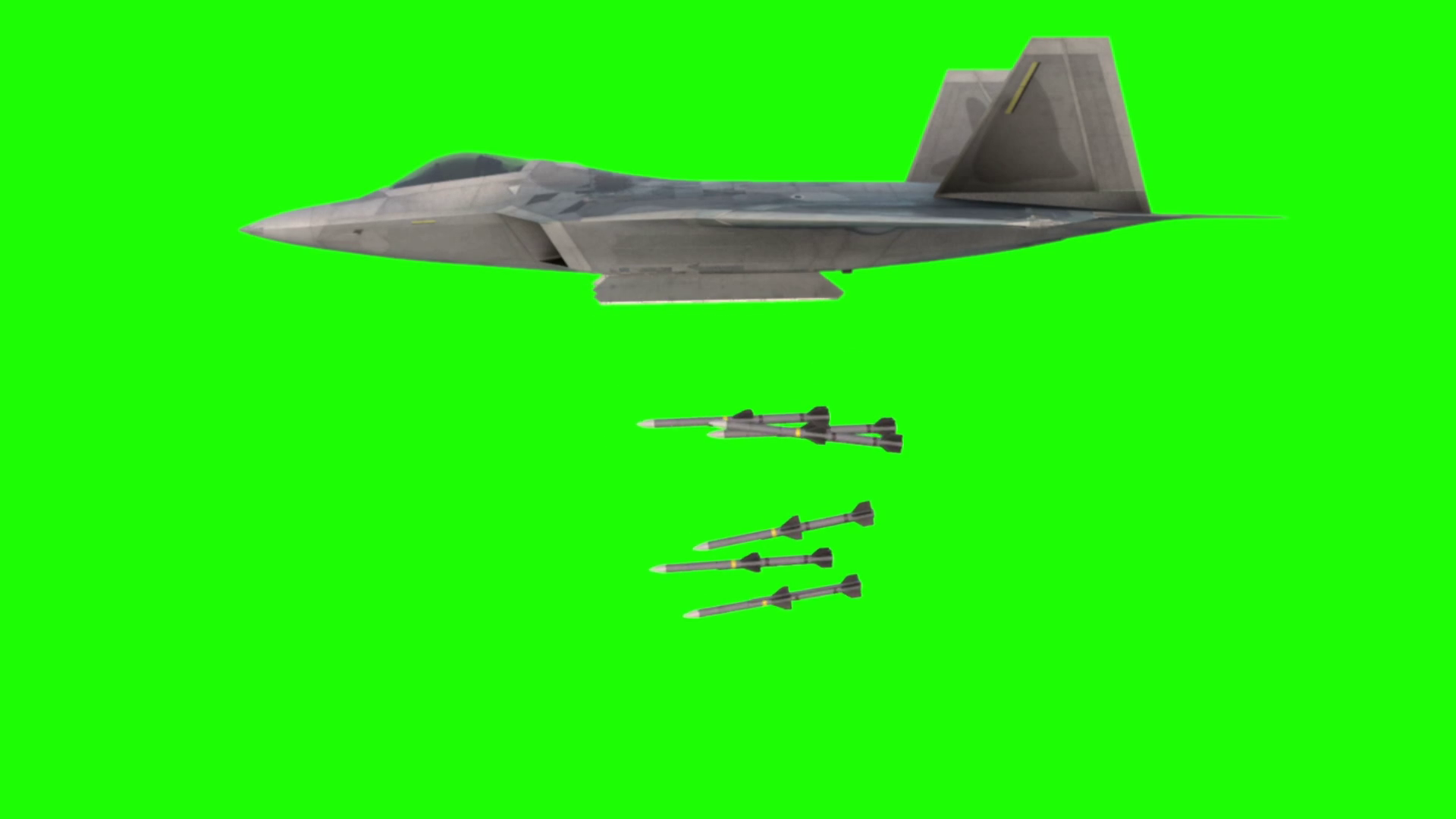}
			\caption{\label{input_green_screen1}Input 1 for green screen removal}
   \end{minipage} \hfill
   \begin{minipage}{.45\linewidth}
   \includegraphics[width=1.\linewidth]{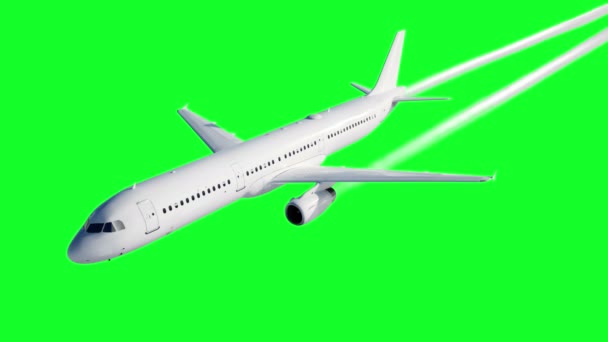}
   \caption{\label{input_green_screen2}Input 2 for green screen removal}
   \end{minipage}
\end{figure}

\begin{figure}[!ht]
   \begin{minipage}{.45\linewidth}
			\caption{\label{coma05_1}Green screen removal with an additional parameter 0.1 (example 1)}
			\includegraphics[scale=0.049]{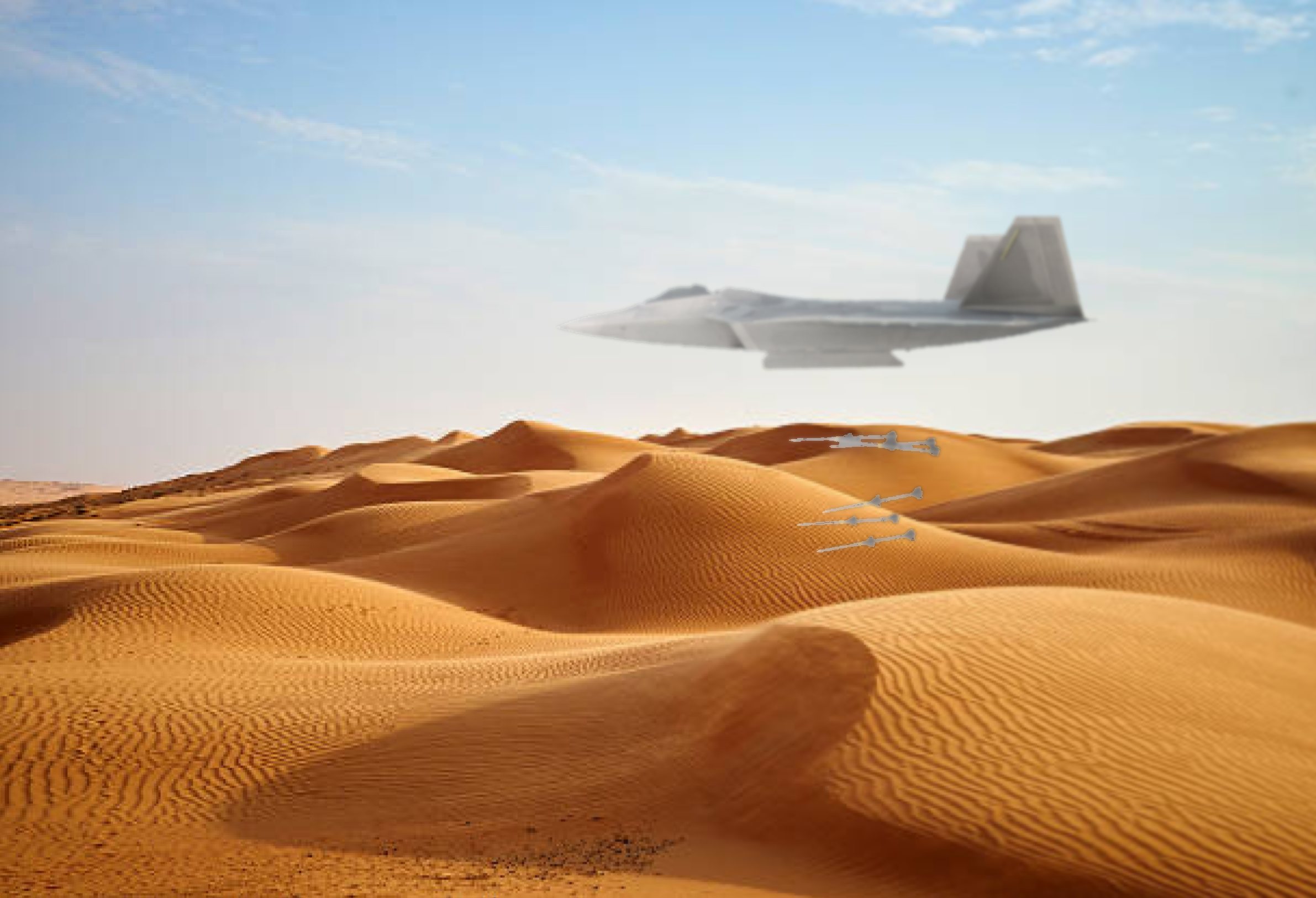}
   \end{minipage} \hfill
   \begin{minipage}{.45\linewidth}
			\caption{\label{coma05_2}Green screen removal with an additional parameter 0.1 (example 2)}
   \includegraphics[scale=0.049]{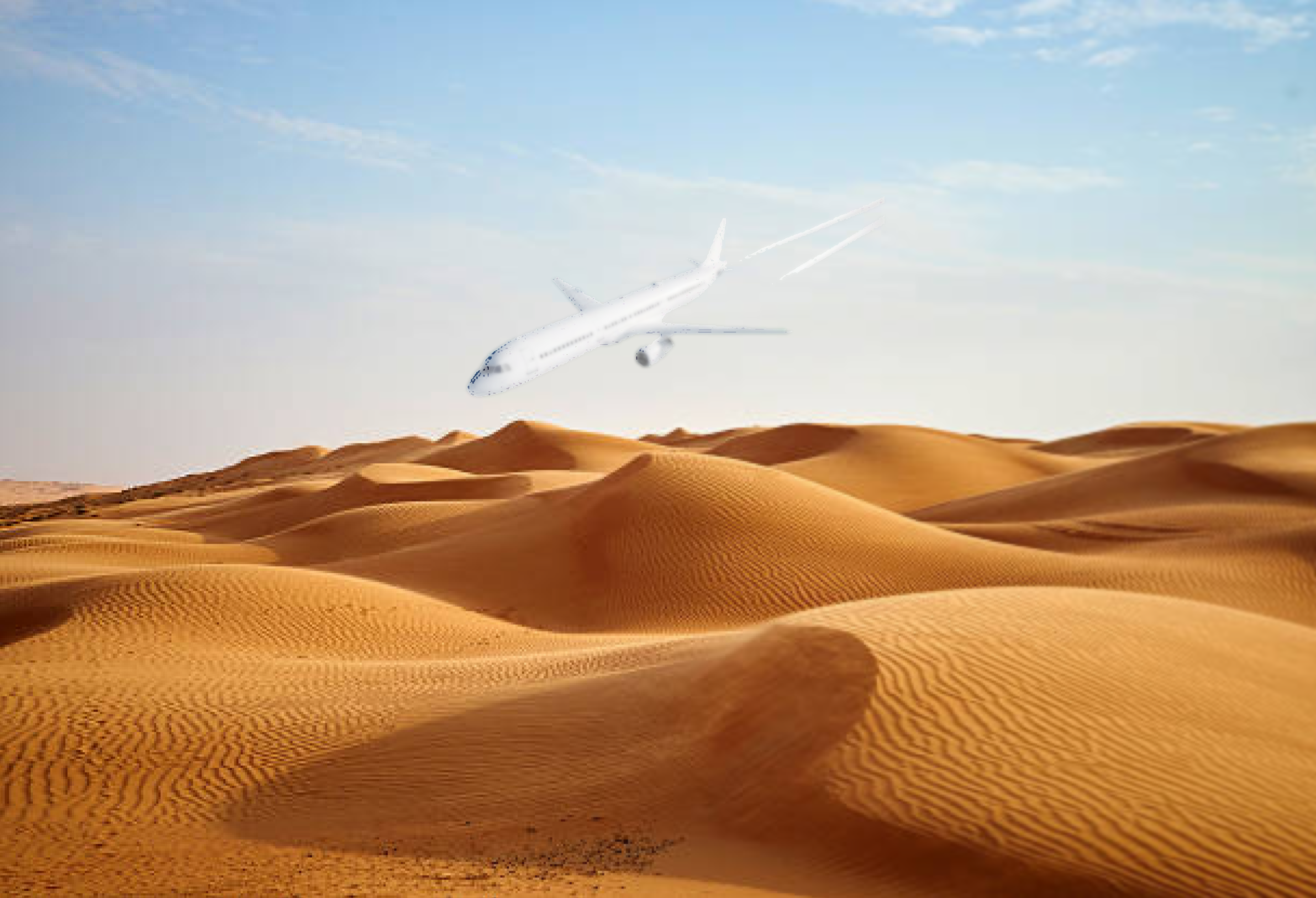}
   \end{minipage}
\end{figure}

\begin{figure}[!ht]
   \begin{minipage}{.45\linewidth}
			\caption{\label{coma20_1}Green screen removal with an additional parameter 0.2 (example 1)}
			\includegraphics[scale=0.049]{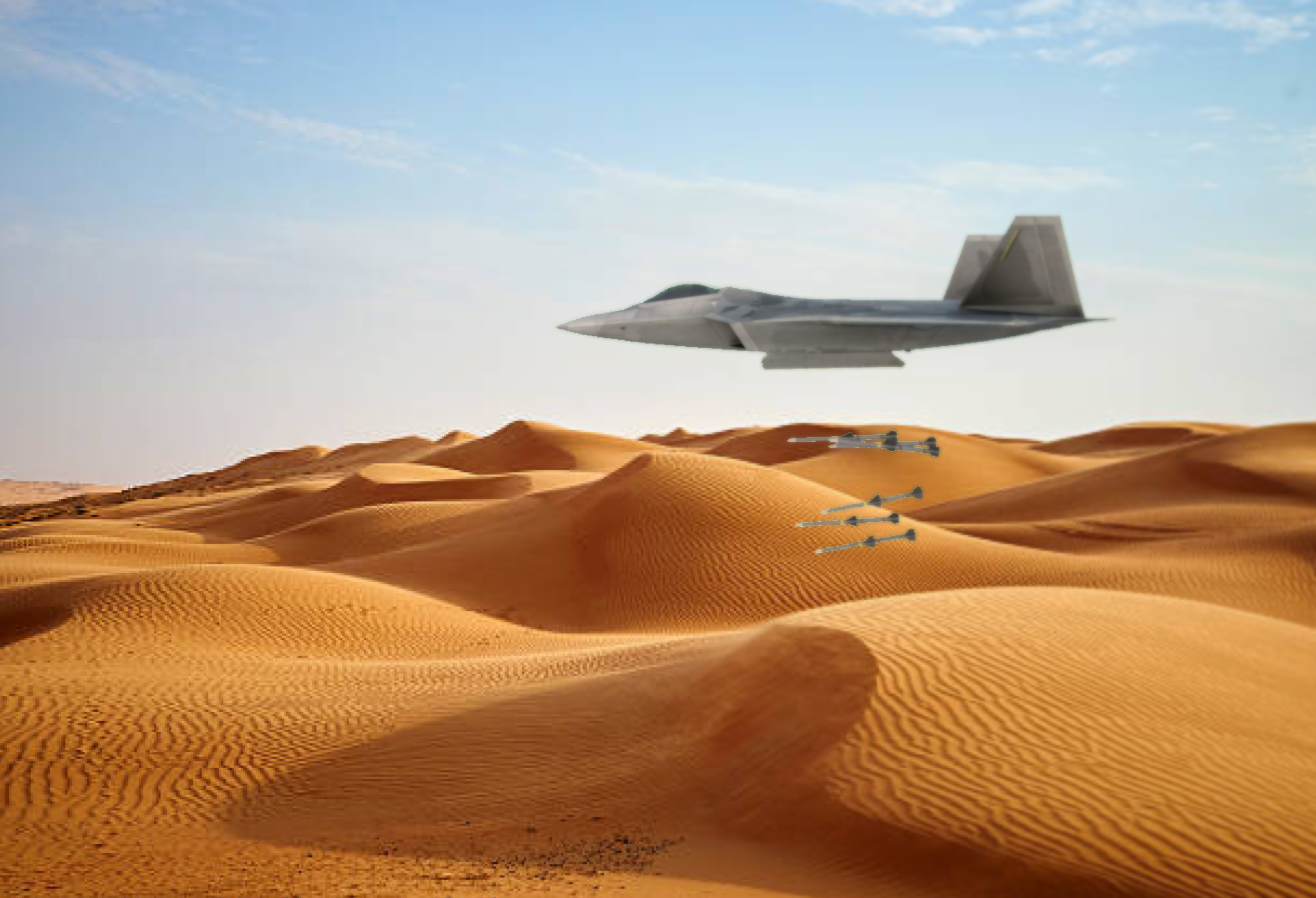}
   \end{minipage} \hfill
   \begin{minipage}{.45\linewidth}
			\caption{\label{coma20_2}Green screen removal with an additional parameter 0.2 (example 2)}
   \includegraphics[scale=0.049]{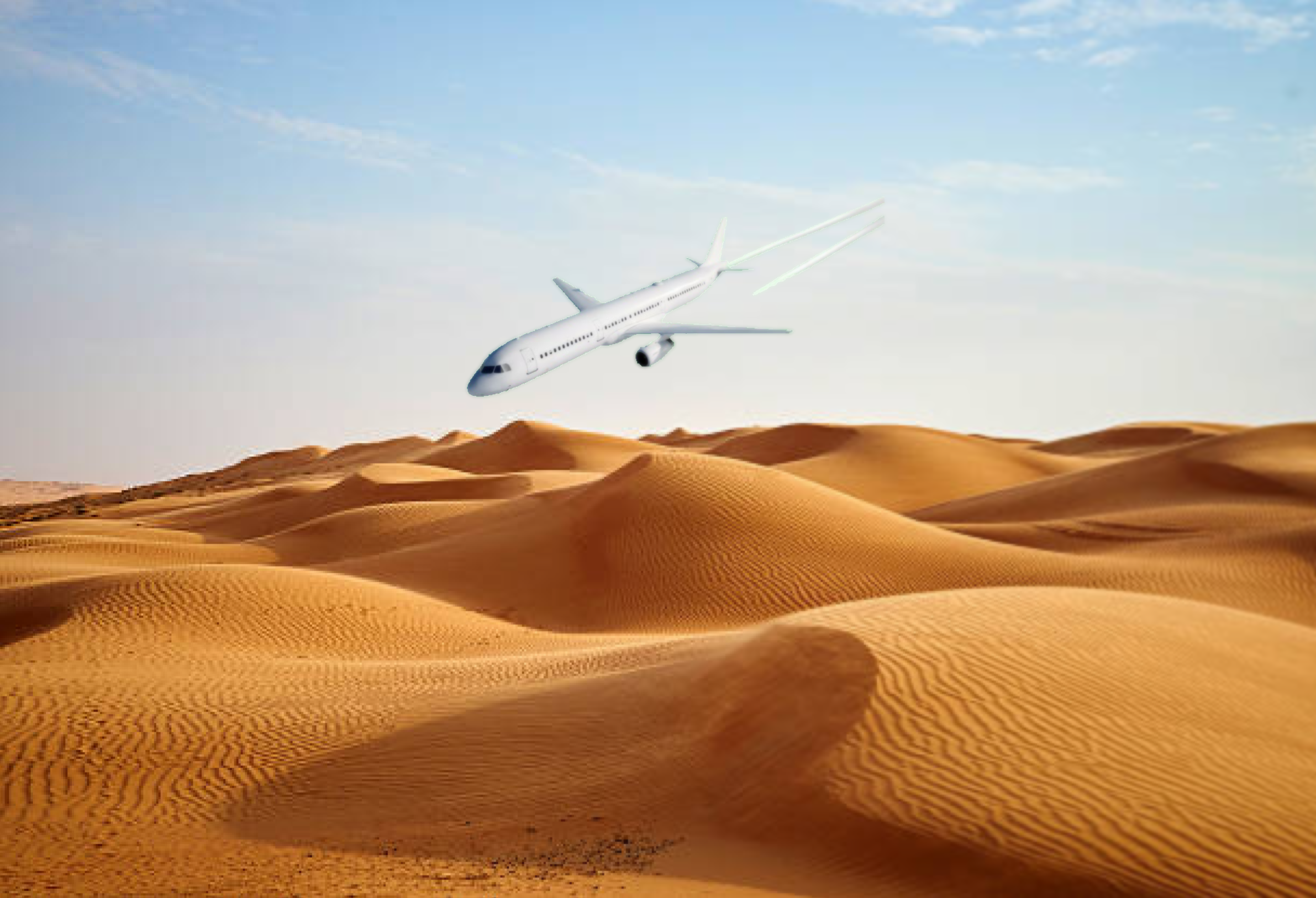}
   \end{minipage}
\end{figure}

\begin{figure}[!ht]
   \begin{minipage}{.45\linewidth}
			\caption{\label{coma50_1}Green screen removal with an additional parameter 0.3 (example 1)}
			\includegraphics[scale=0.049]{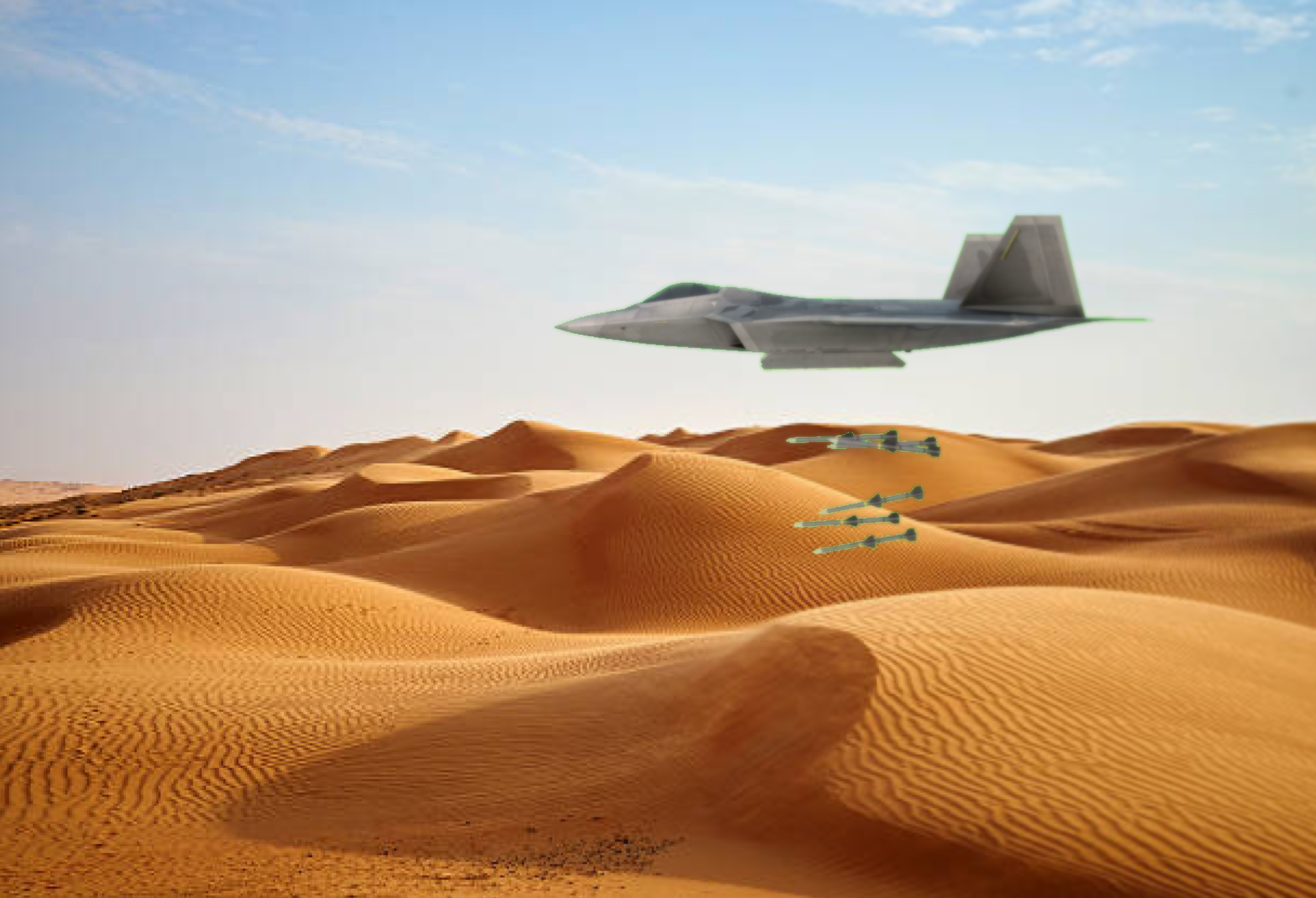}
   \end{minipage} \hfill
   \begin{minipage}{.45\linewidth}
			\caption{\label{coma50_2}Green screen removal with an additional parameter 0.3 (example 2)}
   \includegraphics[scale=0.049]{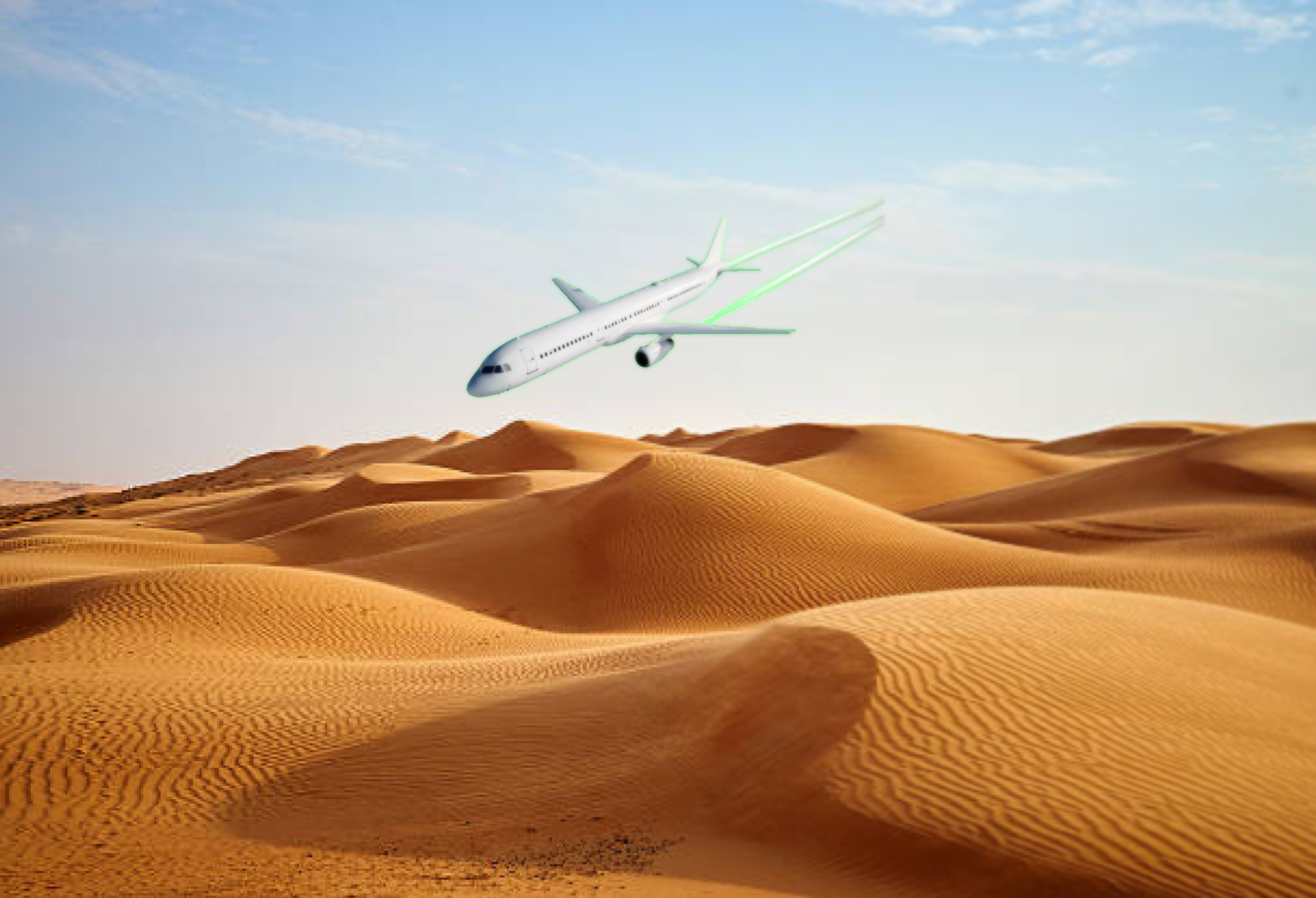}
   \end{minipage}
\end{figure}

In the following pictures (from Figure~\ref{coma05_1} to Figure~\ref{coma50_2}), we can see a comparison of the green screen background removal outputs with this additional parameter varying between $0.1$ and $0.3$ with Figure~\ref{input_green_screen1} and Figure~\ref{input_green_screen2} as inputs with a green screen.

Consequently, when the additional parameter is below $0.2$, some information of the plane is missing. Nevertheless, with values above $0.2$, some green pixels are not correctly removed, and hence the choice of the additional parameter with a value of $0.2$.

\vskip 7pt

The remaining negative values (even after adding $0.2$ to the parameter) become zero in Equation~\eqref{eq:dark}.

\vskip 7pt

After that, the red/blue vs green ratios need to be combined to set an $\alpha$-layer:

\begin{equation}
\alpha \equiv ((blue)vs(green)+(red)vs(green))*255
\label{eq:alpha_layer}
\end{equation}

Then the values of alpha above $50$ become the norm factor value, which is $255$.

\vskip 7pt

Finally, the fourth value $\alpha$ of the NumPy array $RGB\alpha$ is obtained and thus, the new image after setting these tuned parameters is the image without the green background.

\label{green_remov}

\subsubsection{Blending}

After removing the green background, the integration of the two following images must be performed:

\begin{itemize}
	\item The image on the foreground, containing the objects of interest (output of the image segmentation), needs to be an image with an alpha layer (as calculated in the previous Subsection~\ref{green_remov}).
	\item The desired background, which is an inhomogeneous cluttered background.
\end{itemize} 

Furthermore, basic rotation and scaling operations can be done (randomly or in a predefined way) thanks to the \textit{skimage} Python toolkit.

\subsubsection{Channel adjustment}

In order to emphasise colours or other features in an image, a curve is applied to remap the image tonality.

\vskip 7pt

It can be used to each channel individually in an image or to all channels together. The first option can be employed to stress the colour.

\vskip 7pt

The sigmoid function is used to make adjust-curves:

\begin{equation}
S(x)=\frac{1}{1+\e^{-x}}=\frac{\e^{x}}{\e^{x}+1}
\label{eq:sigmoid}
\end{equation}

where $x$ is a NumPy array and $\e^{x}$ is the NumPy exponential function.

\subsubsection{Filtering}

First, the left side of the image is blurred with a Gaussian filter~\cite{BLUR}.

\vskip 7pt

Then, the $\alpha$-values need to be decreased gradually from right to left.

\vskip 7pt

After that, the blurred image and the background are merged into one image to create an appearance of partial or full transparency. This method is called \textit{alpha compositing}~\cite{COMPOS}. The final output of this technique, called the \textit{composite}, is obtained after rendering image elements in separate passes and then combining the resulting multiple 2D images into a single one.

\vskip 7pt

At last, the final outputs of this whole process are saved in a file named \textit{output}, ready to be part of the training dataset.

\subsection{Seg-CK}

\begin{figure}[!ht]
\centering
{\includegraphics[width=0.95\linewidth]{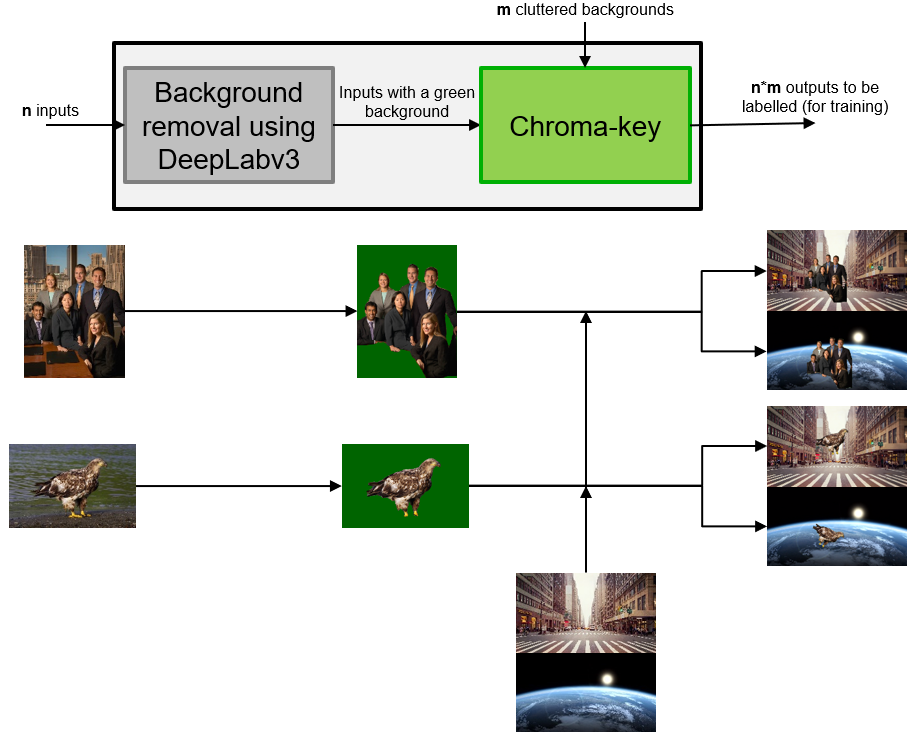}}
\caption{Seg-CK architecture}
\label{methodology}
\end{figure}

The two techniques described in Section~\ref{seg} and Section~\ref{chr}, the image segmentation and Chroma-key, are then integrated as one technique that we named \textbf{Seg-CK}. Its architecture is illustrated in Figure~\ref{methodology}.

Thus, $n$ image inputs first go through the DeepLab background removal process to obtain $n$ outputs with a green background. 

\vskip 7pt

Then, the Chroma-key technique takes two inputs:

\begin{itemize}
	\item The $n$ outputs of the background removal block with a green screen background.
	\item The $m$ relevant heterogeneous backgrounds that can be useful for applications such as fully-autonomous UAV (e.g. backgrounds with a lot of buildings, cars, street furniture, and so on).
\end{itemize}

\vskip 7pt

As a result, we get $n \times m$ outputs with the $n$ images containing the objects of interest that are most likely to be found during a UAV trajectory (e.g. birds, drones, buildings, etc.) in the $m$ cluttered backgrounds.

\subsection{Model training}

\label{training}

\subsubsection{Labelling}

The $n \times m$ outputs are labelled using a GitHub repository named \textit{LabelImg}~\cite{labelimg}.

\vskip 7pt

It is a graphical image annotation tool using Qt for the graphical interface and written in Python.

\begin{figure}[!ht]
\centering
{\includegraphics[width=0.73\linewidth]{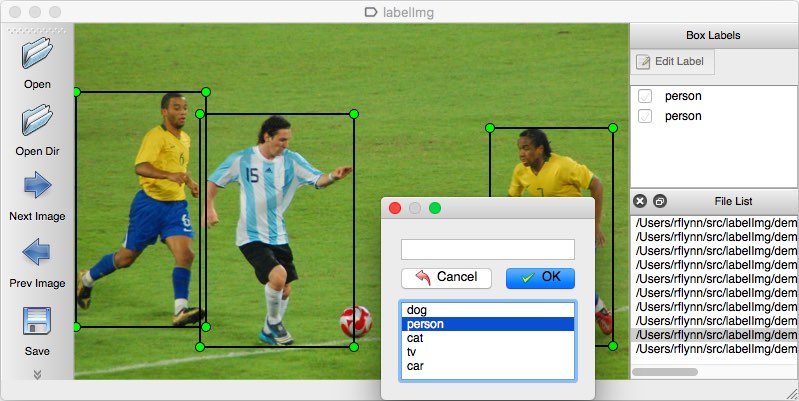}}
\caption{LabelImg for labelling images~\cite{labelimg}}
\label{labelimg}
\end{figure}

\vskip 7pt

The output annotations can be saved in two different formats:

\begin{itemize}
	\item XML files in PASCAL VOC format, used by ImageNet~\cite{IMAGENET}.
	\item TXT files supported by YOLO~\cite{YOLO}.
\end{itemize}

YOLOv3~\cite{YOLOV3} in TensorFlow is used to train the CNN model and to test the performances, hence the TXT format is used. It is a text file containing one or several lines with this specific order: \textit{class} \textit{$x_{min}$} \textit{$y_{min}$} \textit{$x_{max}$} \textit{$y_{max}$}.

\vskip 7pt

For example, if we label a picture containing two classes (bird and aeroplane), we can have:

\vskip 5pt

bird 41 224 224 341
\vskip 2pt
aeroplane 295 80 583 294

\subsubsection{Training}

Two different training dataset is compared for the performance assessment purposes:

\begin{itemize}
	\item The COCO dataset with only its aeroplane class.
	\item The artificial dataset generated from the proposed framework, also with only one aeroplane class.
\end{itemize}

\subsubsection{Training of the COCO dataset}

After training the model with only the aeroplane class from the COCO dataset in darknet~\cite{darknet13}, we just needed to convert the darknet weights to a TensorFlow checkpoint. These convolution weights, with 53 convolutional layers, and therefore called Darknet-53, are trained on ImageNet~\cite{IMAGENET}.

\begin{figure}[!ht]
\centering
{\includegraphics[width=0.43\linewidth]{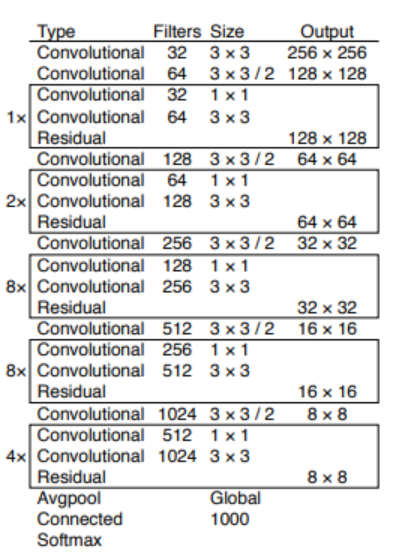}}
\caption{Darknet-53~\cite{YOLOV3}}
\label{darknet53}
\end{figure}

Then, a Python weight converter was used to convert this \textit{yolov3.weights} the darknet weights~\cite{darknet13} to a TensorFlow checkpoint file named \textit{yolov3.ckpt}.

\subsubsection{Training of the dataset created with the Seg-CK framework}

We suppose that we created $10000$ images with the Seg-CK method ($2000$ images of an aeroplane in $5$ different cluttered backgrounds) and that these pictures have been labelled. Also, we wanted to have the same proportion of planes and drones as in the aeroplane class from the COCO dataset.

\vskip 7pt

In order to train a YOLOv3 model with the specific dataset, three main files need to be created:

\begin{enumerate}
	\item \textit{aeroplane.data}
	\item \textit{aeroplane.names}
	\item \textit{aeroplane.cfg}
\end{enumerate}

Firstly, in the \textit{data} file, the details that need to be mentioned are:

\begin{itemize}
	\item The number of classes.
	\item The train set file.
	\item The validation set file.
	\item The file that contains the names of the classes we want to train and hence to detect.
	\item The folder where the yolo weights file is stored.
\end{itemize}

Consequently, we created the following \textit{data} file:

\begin{figure}[!ht]
\centering
{\includegraphics[width=0.6\linewidth]{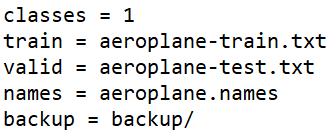}}
\caption{\textit{aeroplane.data} file}
\label{data_file}
\end{figure}

Then, as its name suggests, \textit{aeroplane.names} contains all the names of the classes.

\begin{figure}[!ht]
\centering
{\includegraphics[width=0.6\linewidth]{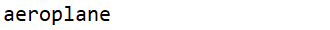}}
\caption{\textit{aeroplane.names} file}
\label{aeroplane_class}
\end{figure}

Last but not least, many vital parameters need to be defined in \textit{aeroplane.cfg}:

\begin{itemize}
	\item $batch = 48$, which means that for each training step, 48 images are used.
	\item $subdivisions = 16$, which is used to decrease the GPU VRAM (Graphics Processing Unit Video Random Access Memory) requirements by dividing the batch by 16.
	\item $classes = 1$, because the only category we train is the aeroplane class.
	\item $filters = (classes + 5)\times 3$, therefore $filters=18$.
\end{itemize}

The batch and subdivisions are tuned parameters to get better performance.

\vskip 7pt

\section{PERFORMANCE ASSESSMENT}

After the Neural Network model was trained, accuracy metrics such as the \textit{mean average precision} (mAP) are used to assess the performances of:

\begin{itemize}
	\item The detection accuracy.
	\item The classification accuracy.
\end{itemize}

\vskip 3pt

We did not conduct an assessment with planes/drones or backgrounds pictures that were trained in the aeroplane class.

\vskip 7pt

However, we chose to train 2000 different images from the aeroplane class in five different cluttered backgrounds ($2000 \times 5 = 10000$ pictures). Different heterogeneous backgrounds were also used (e.g. complex urban environments). 

\vskip 7pt

Nevertheless, as explained in the three following sections, we still get excellent performances in pictures and videos of an aeroplane in other kinds of backgrounds as shown in Figure~\ref{detection_accuracy} and Figure~\ref{3heterogeneous} (e.g. forest, very cloudy sky, buildings, space).

\subsection{Detection}

In this section, detection accuracy is compared between two different models:

\begin{itemize}
	\item The model trained with COCO dataset.
	\item The model trained with the artificial dataset created thanks to the proposed framework.
\end{itemize}

\begin{figure}[!ht]
\centering
{\includegraphics[width=0.83\linewidth]{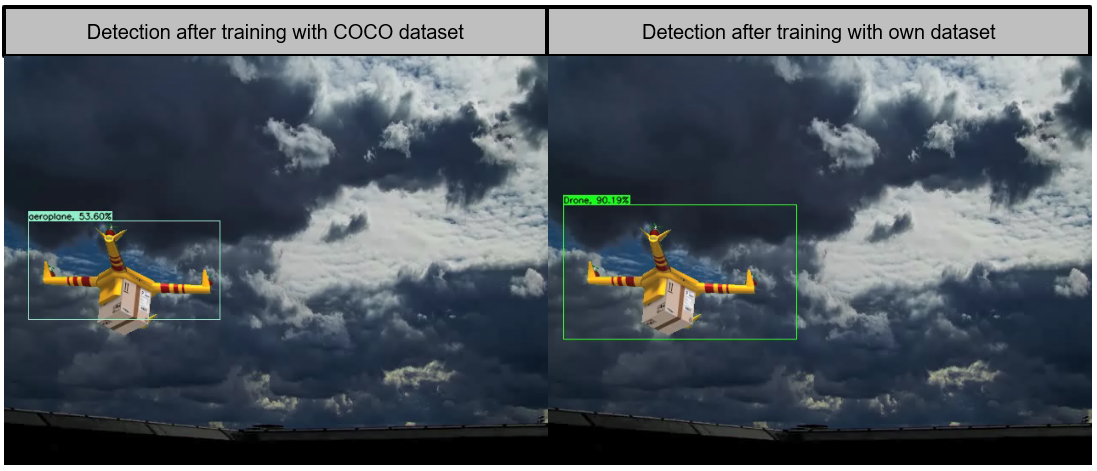}}
\caption{Frame captures of a video applying YOLOv3 in a heterogeneous environment}
\label{detection_accuracy}
\end{figure}

In order to conduct a performance assessment in highly cluttered backgrounds, existing videos of aircraft in heterogeneous environments were taken (e.g. from YouTube).

\vskip 7pt

With the chosen accuracy metric (mAP), only pictures can be evaluated. Hence, the following steps were done:

\begin{enumerate}

\item It is necessary to have two videos: an original video and the same video outputted from YOLO. 

\item Then these two videos need to be divided into the same number of frames, for example, 100 frames. 

\item From the frames of the original video, ground truths are created. 

\item Then, each frame of the ground truth and YOLO result is compared one by one with the mAP metric.

\end{enumerate}

With three videos of aeroplane, which can be split into 300 frames for a total 27 seconds of videos, in heterogeneous backgrounds (such as the one in Figure~\ref{detection_accuracy}), the mAP was approximately $38\%$ for the model trained with the COCO dataset, whereas the average for the model trained with the proposed framework was $86\%$.

\vskip 7pt

These values were auspicious, and thus, classification accuracy needed to be evaluated with a more diversified validation dataset to confirm this analysis.

\subsection{Classification}

In this section, $700$ images of aeroplane are tested in different heterogeneous cluttered backgrounds (images containing planes/drones and backgrounds utterly different from the training set). 

\begin{figure}[!ht]
   \begin{minipage}{.45\linewidth}
			\caption{\label{detection-results-coco}Detection results in complex environments (model trained with the COCO dataset)}
			\includegraphics[scale=0.25]{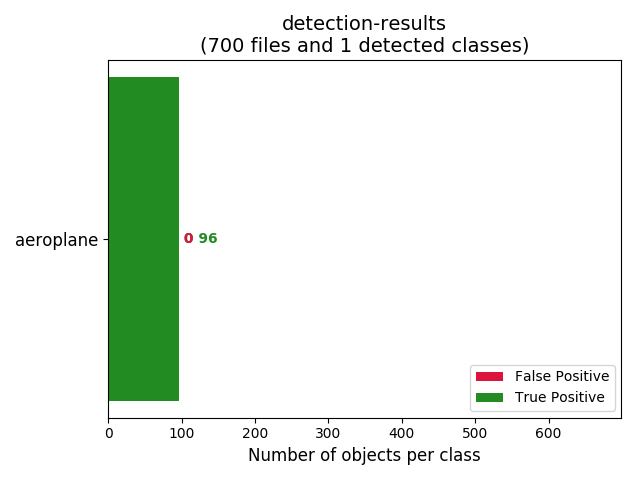}
   \end{minipage} \hfill
   \begin{minipage}{.45\linewidth}
			\caption{\label{detection-results-clutter}Detection results in complex environments (model trained with the proposed framework)}
   \includegraphics[scale=0.25]{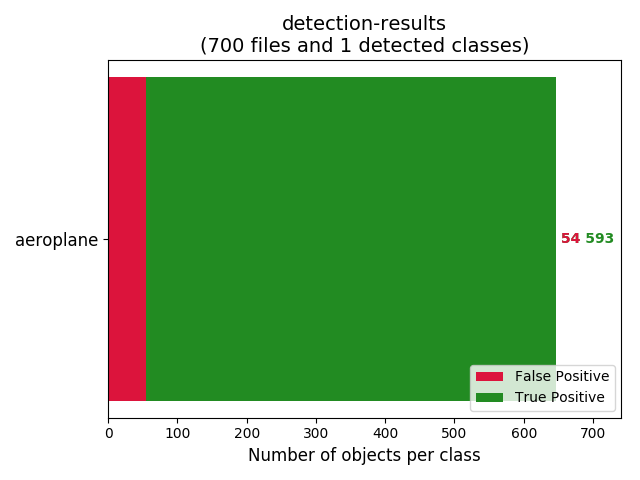}
   \end{minipage}
\end{figure}

Here, on the one hand, the model trained with the COCO dataset's aeroplane class gives a mean average precision of $\frac{96}{700}=13.71\%$, which is an appalling performance. However, the proposed methodology gives a mean average precision of $83.21\%$. In fact, after training the Neural Network model with the proposed framework, YOLOv3 detected 647 times out of 700 that an aeroplane was present, but only 593 times of them were true positives (which means that the prediction was above the predefined threshold IoU of $0.5$). The 54 others are false positives (the IoU is smaller than $0.5$, or a bounding box is duplicated).

\vskip 7pt

Intersection over Union (IoU) represents the overlapping area over the combined area of two bounding boxes. 

\begin{figure}[!ht]
\centering
{\includegraphics[width=0.43\linewidth]{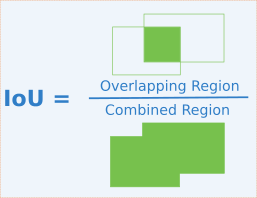}}
\caption{Intersection over Union~\cite{IoU_images}}
\label{iou}
\end{figure}

\newpage

To obtain the IoU, two types of bounding boxes are needed:

\begin{itemize}
	\item The \textit{predicted bounding boxes} from the YOLO model.
	\item The \textit{ground-truth bounding boxes}, which are usually created by the user for comparison means (e.g. by labelling images). 
\end{itemize}

\vskip 7pt

Therefore, an IoU threshold of $0.5$ was predefined since a similar threshold was used for the official COCO~\cite{COCO} dataset accuracy testing. Thus, a true positive is when a label of a picture has an IoU greater than $50\%$ when comparing the predicted bounding box and the ground-truth bounding box. 

\begin{figure}[!ht]
   \begin{minipage}{.45\linewidth}
			\caption{\label{mAP_coco}Mean average precision with complex test dataset (model trained with the COCO dataset)}
			\includegraphics[scale=0.25]{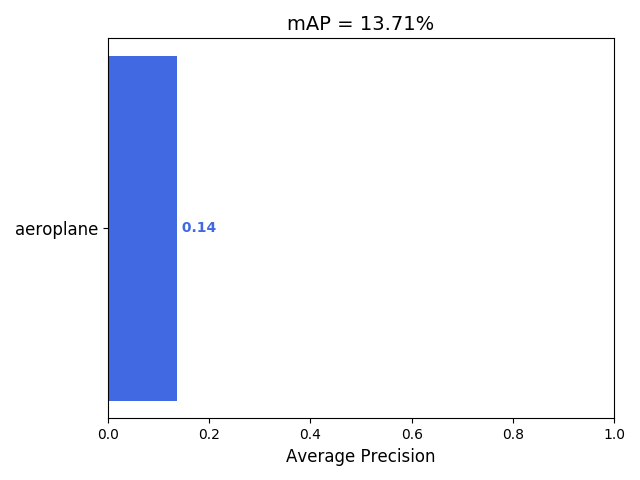}
   \end{minipage} \hfill
   \begin{minipage}{.45\linewidth}
			\caption{\label{mAP_clutter}Mean average precision with complex test dataset (model trained with the proposed framework)}
   \includegraphics[scale=0.25]{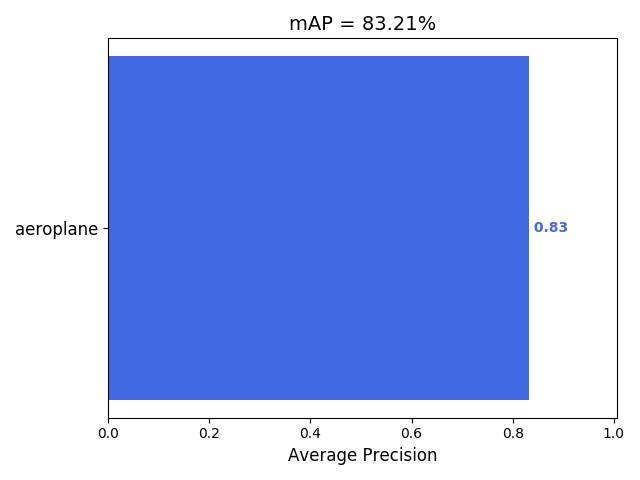}
   \end{minipage}
\end{figure}

The average precision is computed by:

\begin{equation}
AP_{k}= \frac{1}{G_{TP}}\sum_{i=1}^k{\frac{TP_{seen}}{i}}
\label{eq:AP}
\end{equation}

where $G_{TP}$ is the total number of ground-truth positives (labelled-as-positive data), $TP_{seen}$ is the number of true positives seen and $AP_{k}$ is the $k^{th}$ picture's average precision.
\label{AP_IoU_etc}

\vskip 7pt

Finally, the mean of all the Average Precision is considered as the final accuracy metric (mAP):

\begin{equation}
mAP= \frac{1}{N}\sum_{i=1}^N{AP_{i}}
\label{eq:mAP}
\end{equation}

This metric was chosen mostly because it is the principal metric when conducting a performance assessment of a state-of-the-art object detection technique.

\begin{figure}[!ht]
   \begin{minipage}{.45\linewidth}
			\caption{\label{lamr_coco}Log-average miss rate with complex environments (model trained with the COCO dataset)}
			\includegraphics[scale=0.25]{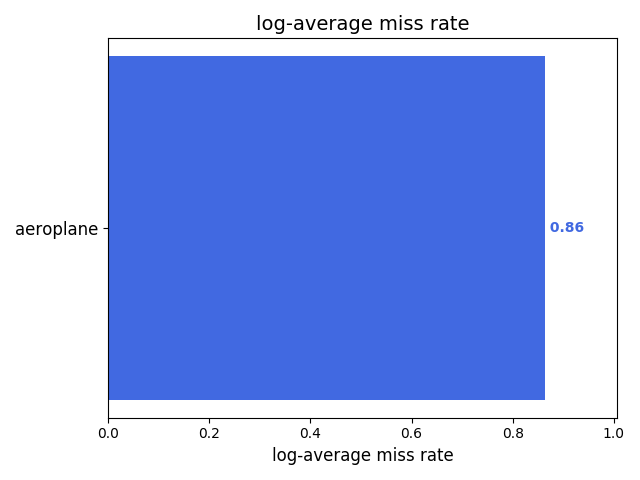}
   \end{minipage} \hfill
   \begin{minipage}{.45\linewidth}
			\caption{\label{lamr_clutter}Log-average miss rate with complex environments (model trained with the proposed framework)}
   \includegraphics[scale=0.25]{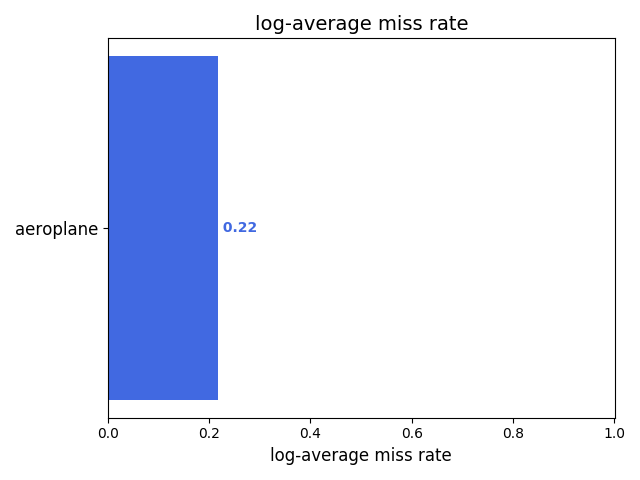}
   \end{minipage}
\end{figure}

The definition of miss-rate is:

\begin{equation}
MR= \frac{FN}{TP+FN}
\label{eq:miss-rate}
\end{equation}

where MR is the miss-rate, TP is the number of true positives and FN is the number of false negatives (which means the IoU is higher than $0.5$, but with a wrong classification).

\begin{figure}[!ht]
   \begin{minipage}{.45\linewidth}
			\caption{\label{aeroplane_coco}Average precision score, micro-averaged over the aeroplane class (model trained with the COCO dataset and tested with complex images)}
			\includegraphics[scale=0.25]{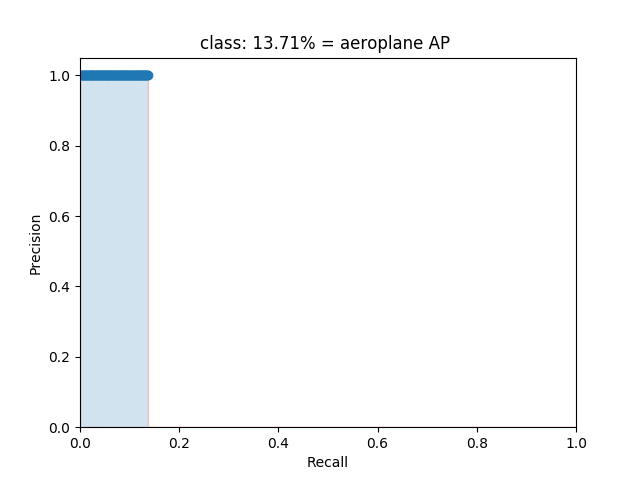}
   \end{minipage} \hfill
   \begin{minipage}{.45\linewidth}
			\caption{\label{aeroplane_clutter}Average precision score, micro-averaged over the aeroplane class (model trained with the proposed framework and tested with complex images)}
   \includegraphics[scale=0.25]{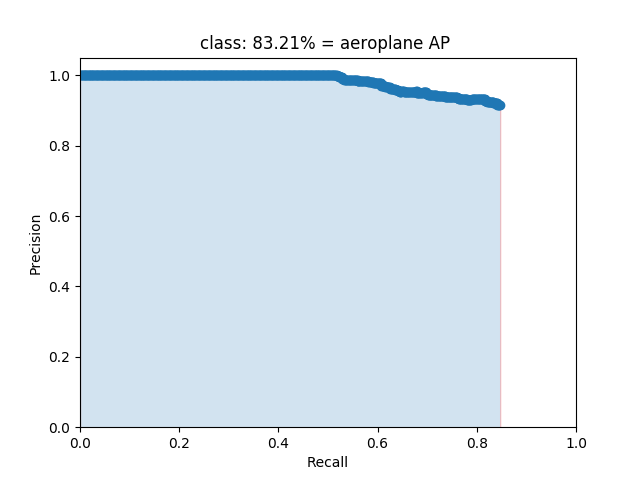}
   \end{minipage}
\end{figure}

Thus, the performance difference is huge. The model trained with the proposed framework gives an improvement of $\frac{83.21}{13.71}=\textbf{607\%}$ compared to the model trained with the COCO dataset when performing an object detection technique (YOLO) in cluttered backgrounds, for one trained class.

\vskip 7pt

Furthermore, COCO dataset is an excellent object detection dataset with 80 classes, $80000$ training images and $40000$ validation images. However, it has $3083$ for the aeroplane class. In terms of comparison, the proposed methodology has $10000$ training images for the aeroplane class (planes and drones with the same proportion as the COCO dataset), including $2000$ different images in five cluttered backgrounds, and $700$ validation images.

\vskip 7pt

Figure~\ref{3heterogeneous} contains three examples of the same UAV in three different heterogeneous backgrounds during a one-by-one detection/classification accuracy assessment in different positions:

\begin{figure}[!ht]
\centering
{\includegraphics[width=1\linewidth]{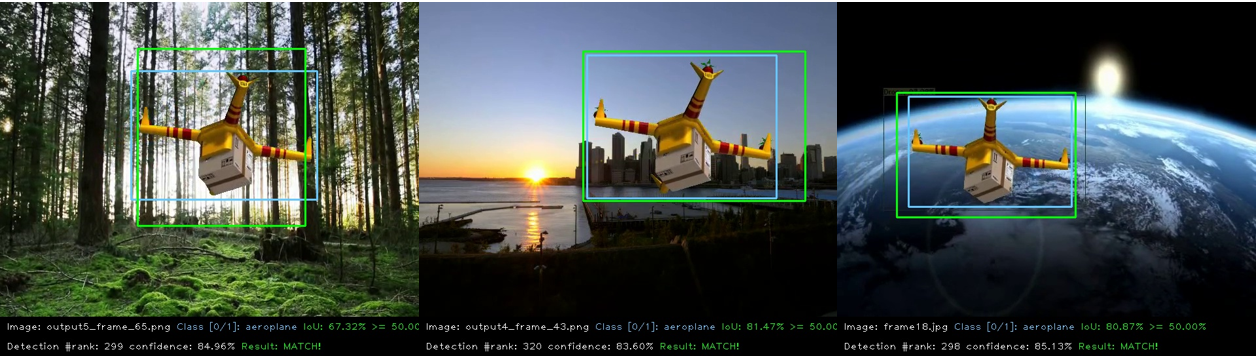}}
\caption{Classification accuracy assessment (mAP) in three different cluttered backgrounds (ground-truth box in blue and YOLOv3 detection result box with the proposed method of training in green) by performing IoU}
\label{3heterogeneous}
\end{figure}

\section{CONCLUSIONS}

Accuracy is an essential requirement in computer vision applications, however background clutter in real-world conditions degrades accuracy performances. In this paper, we focused on improving accuracy when performing a real-time object detection technique called YOLO in highly cluttered environments. For that, an innovative framework of generating an artificial training dataset for neural network models was implemented. It is based on two open-source codes: a state-of-the-art semantic segmentation model DeepLab to extract the objects of interest from the selected images, and Chroma-key, a technique which merges the extracted objects into predefined heterogeneous background. The resulting framework was called Seg-CK and permitted the training process to be more efficient. Moreover, a model trained with the proposed framework was shown to be six times more accurate in cluttered backgrounds than models trained with existing images (e.g. COCO dataset). 

\vskip 7pt

In general, the proposed methodology is shown to be an up-and-coming tool in considered contexts. Therefore, further research on this kind of training method is recommended. For instance, the number of different cluttered backgrounds in the training set is still tunable. Also, further research can be done on GAN (Generative Adversarial Network) in order to generate the backgrounds since it is a potent and promising tool to generate data from scratch.

\addtolength{\textheight}{-12cm}   





\bibliographystyle{unsrt}
\bibliography{library}

\end{document}